\documentclass[sigconf]{acmart}
\AtBeginDocument{%
  \providecommand\BibTeX{{%
    \normalfont B\kern-0.5em{\scshape i\kern-0.25em b}\kern-0.8em\TeX}}}

\copyrightyear{2023} 
\acmYear{2023} 
\setcopyright{acmlicensed}
\acmConference[CIKM '23]{Proceedings of the 32nd ACM International Conference on Information and Knowledge Management}{October 21--25, 2023}{Birmingham, United Kingdom}
\acmBooktitle{Proceedings of the 32nd ACM International Conference on Information and Knowledge Management (CIKM '23), October 21--25, 2023, Birmingham, United Kingdom}
\acmPrice{15.00}
\acmDOI{10.1145/3583780.3615075}
\acmISBN{979-8-4007-0124-5/23/10}

\usepackage{graphicx} 
\graphicspath{{figs/}}
\usepackage{bm}

\DeclareMathOperator*{\argmax}{argmax}

\usepackage{threeparttable}
\usepackage{booktabs}
\usepackage{multicol}
\usepackage{multirow}
\usepackage{booktabs}
\usepackage{enumitem}
\usepackage{hyperref}

\usepackage{xcolor, array, colortbl}

\definecolor{deepred}{rgb}{0.698,0.133,0.133}
\definecolor{blue}{rgb}{0,0,1}

\definecolor{orange}{rgb}{1,0.38,0}
\definecolor{beige}{rgb}{0.639,0.58,0.502}

\definecolor{lightgray}{rgb}{.91,.91,.91}

\begin{document}

\title[Task Relation Distillation and Prototypical Pseudo Label for Incremental Named Entity Recognition]{Task Relation Distillation and Prototypical Pseudo Label \\for Incremental Named Entity Recognition}

\author{Duzhen Zhang}
\authornote{Both authors contributed equally to this research.}
\orcid{0000-0002-4280-431X}
\affiliation{%
  \institution{Baidu Inc.}
  \city{Beijing}
  \country{China}
}
\email{zhangduzhen@baidu.com}

\author{Hongliu Li}
\orcid{0000-0003-2590-9491}
\authornotemark[1]
\affiliation{%
  \institution{The Hong Kong Polytechnic University}
  \city{Hong Kong}
  \country{China}
}
\email{hongliuli1994@gmail.com}

\author{Wei Cong}
\orcid{0000-0002-9531-7179}
\affiliation{%
  \institution{Shenyang Institute of Automation, Chinese Academy of Sciences}
  \city{Shenyang}
  \country{China}
}
\email{congwei45@gmail.com}

\author{Rongtao Xu}
\orcid{0000-0003-4619-9679}
\affiliation{%
  \institution{Beijing Academy of Artificial Intelligence}
  \city{Beijing}
  \country{China}
}
\email{xurongtao2019@ia.ac.cn}

\author{Jiahua Dong}
\orcid{0000-0001-8545-4447}
\affiliation{%
  \institution{Shenyang Institute of Automation, Chinese Academy of Sciences}
  \city{Shenyang}
  \country{China}
}
\email{dongjiahua1995@gmail.com}

\author{Xiuyi Chen}
\authornote{Corresponding author.}
\orcid{0000-0002-9351-4160}
\affiliation{%
  \institution{Baidu Inc.}
  \city{Beijing}
  \country{China}
}
\email{chenxiuyi01@baidu.com}


\begin{abstract}

Incremental Named Entity Recognition (INER) involves the sequential learning of new entity types without accessing the training data of previously learned types. However, INER faces the challenge of catastrophic forgetting specific for incremental learning, further aggravated by background shift (\emph{i.e.,} old and future entity types are labeled as the non-entity type in the current task).
To address these challenges, we propose a method called task Relation Distillation and Prototypical pseudo label (RDP) for INER.
Specifically, to tackle catastrophic forgetting, we introduce a task relation distillation scheme that serves two purposes: 1) ensuring inter-task semantic consistency across different incremental learning tasks by minimizing inter-task relation distillation loss, and 2) enhancing the model's prediction confidence by minimizing intra-task self-entropy loss. Simultaneously, to mitigate background shift, we develop a prototypical pseudo label strategy that distinguishes old entity types from the current non-entity type using the old model.
This strategy generates high-quality pseudo labels by measuring the distances between token embeddings and type-wise prototypes.
We conducted extensive experiments on ten INER settings of three benchmark datasets (\emph{i.e.,} CoNLL2003, I2B2, and OntoNotes5). The results demonstrate that our method achieves significant improvements over the previous state-of-the-art methods, with an average increase of 6.08\% in Micro F1 score and 7.71\% in Macro F1 score.
\end{abstract}







\maketitle

\section{Introduction}

Named Entity Recognition (NER) plays a crucial role in information extraction, benefiting various applications such as question answering~\cite{DBLP:conf/acl/LiYSLYCZL19,DBLP:conf/emnlp/LongprePCRD021}, web search queries~\cite{DBLP:conf/sigir/FetahuFRM21,DBLP:conf/sigir/GuoXCL09,DBLP:conf/aaai/MokhtariMY019,DBLP:conf/kdd/ZhangJD0YCTHWHC21}, and more. NER involves extracting entities from unstructured text and classifying them into predefined entity types (\emph{e.g.,} Person, Location) or the non-entity type.
In the traditional fully-supervised NER paradigm~\cite{neural_for_nlp}, all predefined entity types are learned simultaneously during the training phase, assuming that no new entity types will be encountered during the testing phase. However, real-world applications often involve encountering new entity types on-the-go. For instance, voice assistants like Siri need to recognize new entity types (\emph{e.g.,} Actor, Genre) to understand new intents (\emph{e.g.,} GetMovie)~\cite{monaikul2021continual}. Traditional NER methods struggle to handle new entity types that were not observed during training. One straightforward approach is to fine-tune the NER model using samples of the new entity types. This approach renders existing methods brittle in incremental learning settings, known as Incremental Named Entity Recognition (INER)~\cite{continual_lifelong,lifelong_learning}. There is a need to develop effective INER methods~\cite{monaikul2021continual,xia2022learn,zheng2022distilling} that can incrementally learn the NER model using training samples of novel entity types only. However, INER faces two main challenges: catastrophic forgetting~\cite{catastrophic_1,catastrophic_2,catastrophic_3,catastrophic_4} and background shift.

\begin{figure}[t!]
\centering
  \includegraphics[width=1.0\linewidth]{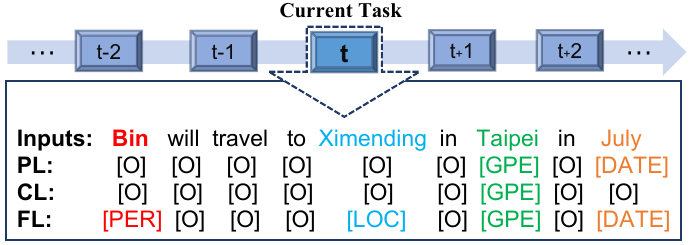} 
    \caption{A simplified INER example, where \textbf{PL}, \textbf{CL}, and \textbf{FL} denote Predicted Labels of the current model, Current ground-truth Labels, and Full ground-truth Labels, respectively. Old entity types (\emph{e.g.,} \textcolor{red}{[PER]} (\texttt{Person}), \textcolor{orange}{[DATE]} (\texttt{Date})) and future entity type (\emph{e.g.,} \textcolor{cyan}{[LOC]} (\texttt{Location})) are labeled as non-entity type (\textcolor{black}{[O]}) in the current task $t$ where \textcolor{green}{[GPE]} (\texttt{Countries, Cities, or States}) is the current entity type being learned, leading to background shift (the third row \textbf{CL}). Furthermore, the NER model incrementally learns new entity types without accessing previous samples, suffering from catastrophic forgetting of old entity types (e.g., the model forgets old entity types \textcolor{red}{[PER]}) (the second row \textbf{PL}).
}
\label{fig:motivation}
\end{figure}

The first challenge, which is inherent in incremental learning, is catastrophic forgetting. This issue arises from the fact that the network weights are adjusted to accommodate the optimal parameter space for newly introduced entity types, without accessing previously encountered entity types. Consequently, the network tends to forget previously learned knowledge after acquiring new information. As depicted in Figure~\ref{fig:motivation} (the second row, \textbf{PL}), after learning the new entity type \texttt{Cities} in the current task, the NER model completely forgets the old entity type \texttt{Person}. Current INER methods~\cite{monaikul2021continual,xia2022learn,zheng2022distilling} employ the widely used knowledge distillation strategy~\cite{KD} to address catastrophic forgetting. However, these methods primarily focus on distilling the output probability distribution for old entity types from the old model to the new model, without considering task relationships. Consequently, they tend to provide the model with excessive stability (\emph{i.e.,} retaining old knowledge) but limited plasticity (\emph{i.e.,} acquiring new knowledge).

The second challenge specific to INER is background shift. In real-world scenarios, NER tasks involve a significant number of non-entity type tokens in training samples (\emph{e.g.,} approximately 89\% of tokens in the OntoNotes5 dataset~\cite{hovy2006ontonotes} belong to the non-entity type), leading to a model bias towards non-entity type~\cite{dice}.
Furthermore, unlike traditional NER methods, where the non-entity type solely consists of tokens that do not belong to any entity type, INER introduces a different labeling paradigm. In INER, both old and future entity types are labeled as non-entity type in the current task. Here, we refer to this phenomenon as the background shift.
As illustrated in Figure~\ref{fig:motivation} (the third row, \textbf{CL}), the old entity types (e.g., \texttt{Person} and \texttt{Date}, have been learned in previous tasks such as $t-1$, $t-2$, etc.) and the future entity type \texttt{Location} (to be learned in future tasks such as $t+1 $, $t+2 $, etc.) are labeled as non-entity type in the current task $t$. If no measures are taken to distinguish tokens belonging to old entity types from the real non-entity type, this background shift phenomenon may exacerbate catastrophic forgetting. Previous methods~\cite{zheng2022distilling} recognize the non-entity type tokens belonging to old entity types to distill causal effect by directly adopting the maximum output probability from the old model.
However, they overlook the prediction errors of the old model, and the recognition results from this method tend to be noisy. Hence, it is crucial to explore a reasonable strategy to correct the errors of the old model and effectively leverage the old entity type information contained within the non-entity type.

 To address the aforementioned challenges, we propose an effective method called task Relation Distillation and Prototypical pseudo label (RDP) for INER.
Firstly, we introduce a task relation distillation scheme that considers task relationships to mitigate catastrophic forgetting. This scheme comprises two components: an inter-task relation distillation loss and an intra-task self-entropy loss, striking a balance between stability and plasticity. 
The inter-task relation distillation loss transfers knowledge from soft labels to the current model's output probabilities. These soft labels are constructed by combining the one-hot ground truth and the output probabilities of the old model, which helps capture the inter-task semantic relations between old tasks and between old and new tasks by smoothing the one-hot ground truth. 
Moreover, the intra-task self-entropy loss enhances the confidence of the current predictions by minimizing self-entropy.
Secondly, we develop a prototypical pseudo label strategy to explicitly retrieve old entity types within the current non-entity type for classification, effectively overcoming the background shift. 
To correct mistaken labels predicted by the old model and produce high-quality pseudo labels, it exploits the distances between token embeddings and type-wise prototypes to reweight the output probabilities of the old model. The code is available at~\href{https://github.com/BladeDancer957/INER_RDP}{\textcolor{blue}{https://github.com/BladeDancer957/INER\_RDP}}.

 In summary, the main contributions of this paper are as follows:

\begin{itemize}
\item We propose a task relation distillation scheme to consider task relationships in different incremental learning tasks, mitigating the catastrophic forgetting problem by constituting a suitable trade-off between stability and plasticity.

\item We introduce a prototypical pseudo label strategy to utilize the old entity type information contained in the non-entity type, better tackling the background shift problem by correcting the prediction error of the old model and producing high-quality pseudo labels. 

\item We conduct extensive experiments on ten INER settings of three benchmark datasets (\emph{i.e.,} CoNLL2003~\cite{sang1837introduction}, I2B2~\cite{murphy2010serving}, and OntoNotes5~\cite{hovy2006ontonotes}). The results demonstrate that our RDP achieves significant improvements over the previous State-Of-The-Art (SOTA) method CFNER~\cite{zheng2022distilling}, with an average gain of 6.08\% in Micro F1 scores and 7.71\% in Macro F1 scores.

\end{itemize}

\section{Related Work}

\subsection{Incremental Learning}

Incremental learning is a learning paradigm that aims to acquire knowledge from continuous tasks while preventing catastrophic forgetting of previously learned tasks~\cite{lifelong_machine,dong2023federated,ICCV2023_HFC}. Incremental learning algorithms can be divided into regularization-based, dynamic architecture-based, and replay-based approaches.
Regularization-based algorithms involve constraining the model weights \cite{EWC, MAS, SI, ONLINE-EWC}, intermediary features~\cite{CLASSIFIER}, or output probabilities~\cite{LWF}.
Dynamic architecture-based algorithms~\cite{PACKNET, DEN, DA} dynamically assign parameters to previous tasks to maintain the stability of the model as it learns new tasks.
Replay-based algorithms~\cite{icarl, GEM, DGR, FCIL} incorporate a few generated or stored old samples into the current training phase to facilitate learning the new task while retaining knowledge from previous tasks.

\subsection{INER} \label{cner}

Traditional NER aims to classify each token in a sentence into a fixed set of predefined entity types (\emph{e.g.,} Person, Location, etc.) or non-entity type~\cite{ma2016end}. It primarily focuses on designing various deep learning networks to improve NER performance in an end-to-end manner~\cite{li2020survey}, such as BiLSTM-CRF~\cite{DBLP:conf/naacl/LampleBSKD16} and BERT~\cite{kenton2019bert} based methods. However, in more realistic scenarios, new entity types appear periodically on demand, and the NER model should be able to continuously identify new entity types without retraining from scratch~\cite{zhang2023decomposing}. To this end, INER integrates an incremental learning paradigm with traditional NER.

Given that INER is a relatively emerging field, only a handful of recent studies have specifically addressed this problem. ExtendNER~\cite{monaikul2021continual} employs knowledge distillation by transferring output probabilities from the teacher model (\emph{i.e.,} the old model) to the student model (\emph{i.e.,} the new model). L\&R~\cite{xia2022learn} introduces a two-stage framework called learn-and-review: the learning stage utilizes knowledge distillation similar to ExtendNER, while the reviewing stage adopts a replay-based approach, augmenting the current training set with synthetic samples of old entity types. CFNER~\cite{zheng2022distilling} achieves SOTA performance by designing a causal framework and distilling causal effects from non-entity type.
Despite the notable progress achieved by these methods, they employ knowledge distillation in a general manner, focusing solely on maintaining consistency in output probabilities for old entity types while disregarding task relationships. This limitation hampers the model's plasticity (\emph{i.e.,} the ability to learn new information). 
Additionally, they do not effectively address the background shift problem. 
CFNER~\cite{zheng2022distilling} is the first to directly identify non-entity type tokens belonging to old entity types based on the highest prediction probability of the old model to distill causal effects. 
However, this approach propagates errors resulting from incorrect predictions of the old model to the new model. While CFNER employs a curriculum learning strategy to mitigate prediction errors, it heavily relies on manually predefined hyper-parameters, thus restricting its applicability and effectiveness in reducing errors.

By contrast, we design a task relation distillation scheme to consider task relationships among different incremental learning tasks, constituting a suitable trade-off between the model's stability and plasticity and effectively alleviating catastrophic forgetting. 
Furthermore, we develop a prototypical pseudo label strategy for classification, which exploits the distances between token embeddings and type-wise prototypes to reweight the prediction probabilities of the old model, better correcting and reducing the prediction errors and handling the background shift problem.

\vspace{-0.05mm}
\section{Preliminary}

In this section, we present the problem formulation of INER. Following previous works ~\cite{monaikul2021continual,xia2022learn,zheng2022distilling}, INER considers learning a model on a sequence of tasks $\mathcal{Z}^1, \mathcal{Z}^2, ..., \mathcal{Z}^T$, where each task has its corresponding dataset $\mathcal{D}^1, \mathcal{D}^2, ..., \mathcal{D}^T$. The current model $\mathcal{M}^t$ is trained only on the current dataset $\mathcal{D}^t$, resulting in catastrophic forgetting of old entity types in previous datasets $\mathcal{D}^1, \mathcal{D}^2, ..., \mathcal{D}^{t-1}$. Specifically, the current dataset $\mathcal{D}^t$ consists of a set of training samples $(X^t, Y^t)$, where $X^t$ denotes an input token sequence of length $|X^t|$ and $\bm{Y}^t$ denotes the corresponding ground truth label sequence in one-hot form.
 The latter only contains the labels of current entity types $\mathcal{E}^t$ specific to the current task $\mathcal{Z}^t$. All other labels (\emph{e.g.,} potential old entity types $\mathcal{E}^{1:t-1}$ or future entity types $\mathcal{E}^{t+1:T}$) are collapsed into the non-entity type $e_{o}$. Labeling old and future entity types as non-entity type $e_o$ leads to the problem of background shift. Note that entity types in different tasks are non-overlapping, \emph{i.e.,} $\mathcal{E}^i \cap \mathcal{E}^j = \emptyset$ if $i \neq j$. Here, $E^t=card(\mathcal{E}^t)$ represents the cardinality of the current entity types. Typically, the current model $\mathcal{M}^t$ consists of an encoder $F^t$ and a linear softmax classifier $C^t$. Given the current dataset $\mathcal{D}^t$ and the old model $\mathcal{M}^{t-1}$, the goal of INER is to learn the current (new) model $\mathcal{M}^t$ to recognize all entity types $\mathcal{E}^{1:t}$ seen so far. The cross-entropy loss in the current task $\mathcal{Z}^t$ is defined as follows:
\begin{equation}
	\mathcal{L}_{\text{ce}}(\Theta^t)=-\frac{1}{|X^t|}\sum_{i=1}^{|X^t|}\bm{Y}^t(i)\log \bm{\widehat{Y}}^t(i),
	\label{cross_entropy_loss}
\end{equation}
where $\bm{Y}^t(i)\in\mathbb{R}^{1+E^1+...+E^t}$ is the one-hot ground truth label for the token $X^t(i)$, $\bm{\widehat{Y}}^t(i)\in\mathbb{R}^{1+E^1+...+E^t}$ is the output probability distribution of the current model for the token $X^t(i)$, and $\Theta^t$ is the set of learnable parameters of the current model~$\mathcal{M}^t$.

\section{Method}

\begin{figure*}[tbp]
  \centering
  \includegraphics[width=1.0\linewidth]{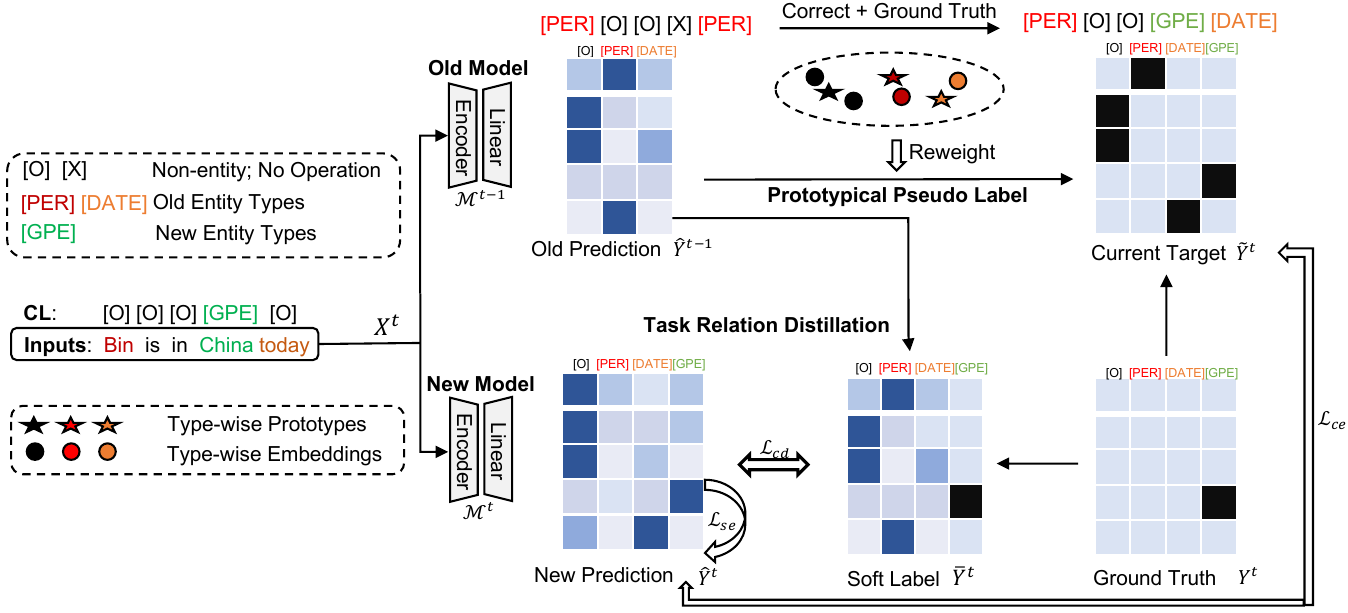} 
   \caption{The overall framework of our RDP, demonstrated by a simplified INER example. \textbf{CL} denotes the current ground-truth labels. For a current input token sequence $X^t$, the soft label $\bm{\overline{Y}}^t$ is calculated by combining the old prediction $\bm{\widehat{Y}}^{t-1}$ with the current ont-hot ground truth $\bm{Y}^t$. The current target $\bm{\widetilde{Y}}^t$ is obtained by the prototypical pseudo label strategy. Then, we update the new model $\mathcal{M}^t$ with the task relation distillation loss (\emph{e.g.,} $\mathcal{L}_{\text{cd}}$ and $\mathcal{L}_{\text{se}}$) and pseudo label based cross entropy loss (\emph{i.e.,} $\mathcal{L}_{\text{ce}})$.
}
   \label{fig:framework}
\end{figure*}

In this section, we introduce our RDP method, designed to facilitate sequential learning of new entity types. To address catastrophic forgetting, we present the task relation distillation scheme in Section~\ref{sec:relation_distillation}. Additionally, we discuss the prototypical pseudo label strategy in Section~\ref{sec:prototype_refinement}, which aims to generate high-quality pseudo labels and tackle the problem of background shift. The overall framework of our proposed RDP method is illustrated in Figure~\ref{fig:framework}.

\subsection{Task Relation Distillation}
\label{sec:relation_distillation}

When learning the current task $\mathcal{Z}^t$, the model $\mathcal{M}^t$ is trained solely on the training samples associated with the current entity types $\mathcal{E}^t$, without any annotations for previously encountered entity types $\mathcal{E}^1, \mathcal{E}^2,..., \mathcal{E}^{t-1}$. This phenomenon often leads to catastrophic forgetting of old entity types. Knowledge distillation~\cite{KD}, a widely adopted technique for retaining prior knowledge, has been demonstrated as effective in previous INER works~\cite{monaikul2021continual,xia2022learn,zheng2022distilling}. These methods alleviate catastrophic forgetting by distilling the output probabilities from the old model $\mathcal{M}^{t-1}$ to the current model $\mathcal{M}^t$. Specifically, the objective function for distilling the output probability distribution can be expressed as:
\begin{equation}
	\mathcal{L}_{\text{kd}}(\Theta^t)=-\frac{1}{|X^t|}\sum_{i=1}^{|X^t|} \bm{\widehat{Y}}^{t-1}(i)\log\bm{\widehat{Y}}^t(i),
    \label{kl_loss}
\end{equation}
where $\bm{\widehat{Y}}^{t-1}(i)\in\mathbb{R}^{1+E^1+...+E^{t-1}}$ and $\bm{\widehat{Y}}^{t}(i)\in\mathbb{R}^{1+E^1+...+E^t}$ represent the output probability distribution of token $X^t(i)$ by the old model $\mathcal{M}^{t-1}$ and the new model $\mathcal{M}^{t}$, respectively (taking the first $1+E^1+...+E^{t-1}$ dimensions of $\bm{\widehat{Y}}^{t}(i)$ to keep the same size with $\bm{\widehat{Y}}^{t-1}(i)$).

However, the current approach to knowledge distillation only focuses on maintaining the consistency of output probabilities for old entity types, disregarding the importance of considering task relationships. This approach tends to provide the model with excessive stability (\emph{i.e.,} retaining old knowledge) while compromising its plasticity (\emph{i.e.,} ability to learn new knowledge). Recent studies suggest that the distillation loss should strike a suitable balance between stability and plasticity~\cite{DBLP:conf/cvpr/DouillardCDC21,incremental_survey}. In light of this, we propose a task relation distillation scheme incorporating an inter-task relation distillation loss (representing stability) and an intra-task self-entropy loss (representing plasticity) to achieve an improved stability-plasticity trade-off for the continual learner.

\paragraph{Inter-task Relation Distillation Loss}

Taking inspiration from human learning mechanisms, which often consider the semantic relationships between old and new categories to enhance the learning of new categories, we aim to distill the underlying semantic relations between old and new entity types, ensuring inter-task semantic consistency across different incremental learning tasks. 
Specifically, we distill knowledge from the constructed soft label $\bm{\overline{Y}}^t\in \mathbb{R}^{|X^t|\times(1+E^1+...+E^t)}$ to the current output probability distribution $\bm{\widehat{Y}}^t\in\mathbb{R}^{|X^t|\times(1+E^1+...+E^t)}$. 
For each token $X^t(i), i=1,...,|X^t|$, $\bm{\overline{Y}}^t(i)\in\mathbb{R}^{1+E^1+...+E^t}$ is obtained by replacing the first $1+E^1+...+E^{t-1}$ dimensions of the one-hot ground truth $\bm{Y}^t(i)\in \mathbb{R}^{1+E^1+...+E^{t}}$ with the old output probability distribution $\bm{\widehat{Y}}^{t-1}(i)\in \mathbb{R}^{1+E^1+...+E^{t-1}}$ (\emph{i.e.,} the first $1+E^1+...+E^{t-1}$ dimensions of $\bm{\overline{Y}}^t(i)$ are $\bm{\widehat{Y}}^{t-1}(i)$, and the last $E^t$ dimensions of $\bm{\overline{Y}}^t(i)$ are $\bm{Y}^t(i)$). By smoothing the one-hot ground truth, the obtained soft label can effectively capture the inter-task semantic relations between old and new entity types. Consequently, the inter-task relation distillation loss $\mathcal{L}_{cd}$ can be formulated as follows:
\begin{equation}
	\mathcal{L}_{\text{cd}}(\Theta^t)=-\frac{1}{|X^t|}\sum_{i=1}^{|X^t|} \bm{\overline{Y}}^t(i)\log\bm{\widehat{Y}}^t(i),
	\label{cd_loss}
\end{equation}

Note that, compared to the vanilla knowledge distillation loss (Equation (\ref{kl_loss})), which can only capture semantic relations between old tasks, our proposed inter-task relation distillation loss (Equation (\ref{cd_loss})) can capture not only semantic relations between old tasks but also capture semantic relations between old tasks (including old entity types) and new tasks (including new entity types).

\paragraph{Intra-task Self-entropy Loss}

After considering the potential semantic relations between old and new categories, humans require making correct choices with high confidence gradually. Self-entropy, known for reflecting the model's confidence, decreases as the model becomes more certain in its predictions. Therefore, we aim to minimize the intra-task self-entropy of the current output probability distribution $\bm{\widehat{Y}}^t$ to enhance the model's confidence. This can be formulated as follows, focusing on increasing the model's certainty in its predictions:
\begin{equation}
	\mathcal{L}_{\text{se}}(\Theta^t)= -\frac{1}{|X^t|}\sum_{i=1}^{|X^t|}\bm{\widehat{Y}}^t(i) \log\bm{\widehat{Y}}^t(i),\\
	\label{sharp_regularizer}
\end{equation}

Overall, the task relation distillation loss is as the following:
\begin{equation}
	\mathcal{L}_{\text{rd}}(\Theta^t)= \lambda_1\mathcal{L}_{\text{cd}}(\Theta^t)+\lambda_2\mathcal{L}_{\text{se}}(\Theta^t) + \mathcal{L}_{\text{kd}}(\Theta^t),\\
	\label{ss_distill}
\end{equation}
where $\lambda_1$ and $\lambda_2$ are the hyper-parameters to balance the stability and plasticity of the NER model.

\subsection{Prototypical Pseudo Label}
\label{sec:prototype_refinement}

As mentioned earlier, the tokens labeled as the non-entity type in the current task $\mathcal{Z}^t$ may belong to old, future, and actual non-entity types. This background shift phenomenon exacerbates the problem of catastrophic forgetting. It is crucial to note that future entity types are unknown and cannot be utilized in the current task. Nevertheless, we can still leverage the old model to identify non-entity type tokens belonging to old entity types.
To address the background shift problem, we employ the pseudo label strategy~\cite{pseudo-label}, commonly used in semantic segmentation domain adaptation approaches~\cite{advent,bidirectional,unsupervised_domain,DBLP:conf/cvpr/Zhang0Z0WW21}. 
However, in our INER setting, instead of assigning pseudo labels to unlabeled target domains, our objective is to utilize the predictions of the old model for non-entity type tokens as clues regarding their actual entity type, particularly if they belong to any of the old entity types.
The most straightforward pseudo label strategy, referred to as the naive pseudo label strategy, involves selecting the most probable old type (including old entity types and the non-entity type) for the non-entity type token predicted by the old model. To this end, we define $\bm{\widetilde{Y}}^t\in\mathbb{R}^{|X^t|\times(1+E^1+...+E^t)}$ as the target for the current task $\mathcal{Z}^t$. This target is calculated with the one-hot ground-truth label sequence $\bm{Y}^t$ and the old output probability distribution $\bm{\widehat{Y}}^{t-1}$ obtained by passing $X^t$ through the old model $\mathcal{M}^{t-1}$, formulated as follows:
\begin{equation}
\bm{\widetilde{Y}}^{t}\left(i,e\right)= \mkern-5mu \left\{\begin{array}{ll}
\mkern-10mu 1 \mkern-27mu &\ \  \text { if } \bm{Y}^{t} (i,e_o)=0 \text { and } e = \argmax \limits_{e' \in \mathcal{E}^{t}} \bm{Y}^{t}(i,e') \\
\mkern-10mu 1 \mkern-27mu &\ \  \text { if } \bm{Y}^{t}(i,e_{o})=1 \text { and } e = \mkern-8mu \argmax \limits_{e' \in e_o\cup\mathcal{E}^{1:t-1}} \mkern-6mu \bm{\widehat{Y}}^{t-1}(i,e') \\
\mkern-10mu 0 \mkern-27mu & \ \ \text { otherwise }\\
\end{array}\right.
\label{eq:pseudo_bis}
\end{equation}
where $\bm{\widetilde{Y}}^t(i,e) = 1$ denotes the pseudo label of the $i$-th token $X^t(i)$ belonging to type $e$ (in practical use, type $e$ needs to be converted to an integer index first). In other words, we copy the ground truth label if tokens are not currently labeled as non-entity type $e_o$ (the first line of Equation~(\ref{eq:pseudo_bis})). Otherwise, we use the label predicted by the old model (the second line of Equation~(\ref{eq:pseudo_bis})).
It is well-known that neural networks often suffer from the problem of over-confidence, leading them to select pseudo labels with high confidence. However, high network confidence does not guarantee the correctness and can prevent the network from learning reliable knowledge. Furthermore, due to the distribution shift among different tasks, this naive pseudo label strategy often introduces noisy labels, negatively impacting the model's performance. 
It is likely that for non-entity type tokens, the closer they are to the distribution of old types, the higher the probability score (\emph{i.e.,} confidence) the old model assigns.

To this end, we propose a prototypical pseudo label strategy to obtain high-quality pseudo labels. We rectify the noisy pseudo labels by reweighting the output probability distribution of the old model according to the calculated type-wise prototypical weights. Equation (\ref{eq:pseudo_bis}) can be modified as follows:
\begin{equation}
\bm{\widetilde{Y}}^{t}\left(i,e\right)= \mkern-5mu \left\{\begin{array}{ll}
\mkern-10mu 1 \mkern-27mu &\ \  \text { if } \bm{Y}^{t} (i,e_o)=0 \text { and } e = \argmax \limits_{e' \in \mathcal{E}^{t}} \bm{Y}^{t}(i,e') \\
\mkern-10mu 1 \mkern-27mu &\ \  \text { if } \bm{Y}^{t}(i,e_{o})=1 \text { and } e = \mkern-8mu \argmax \limits_{e' \in e_o\cup\mathcal{E}^{1:t-1}} \mkern-6mu \bm{\omega}^t(i,e') \bm{\widehat{Y}}^{t-1}(i,e') \\
\mkern-10mu 0 \mkern-27mu & \ \ \text { otherwise }\\
\end{array}\right.
\label{eq:pseudo_proto}
\end{equation}
where $\omega^t(i,e)$ is the type-wise prototypical weight, which measures the distances between the token $X^t(i)$'s embedding and type-wise prototype $\eta^e$, formulated as follows:
\begin{equation}
	\omega^t(i,e) =\frac{\text{exp}(-||F^{t-1}(X^t(i))-{\eta}^{e}||/\tau)}{\sum_{e}\text{exp}(-||F^{t-1}(X^t(i))-{\eta}^{e}||/\tau)},
	\label{prototypocal_weight}
\end{equation}
where $||\cdot||$ denotes the calculation of the L2 norm, $\tau$ is the temperature hyper-parameter, and we empirically set $\tau=1$. $F^{t-1}(X^t(i))$ represents the embedding of the token $X^t(i)$ by the encoder $F^{t-1}$ of the old model $\mathcal{M}^{t-1}$. ${\eta}^{e}$ represents the prototype for type $e$. If the token embedding $F^{t-1}(X^t(i))$ is significantly distant from the prototype ${\eta}^{e}$ (\emph{i.e.,} the feature centroids of type $e$), it is more likely that the learned feature is an outlier. Therefore, we down-weight its probability of being classified into type $e$ and vice versa.

To obtain the prototypes $\eta^e$, we initially obtain the coarse (naive) pseudo labels, which correspond to the indices of the highest output probabilities predicted by the old model for all non-entity type tokens in the current dataset $\mathcal{D}^t$. Subsequently, based on these coarse pseudo labels, we calculate the average embeddings of the tokens predicted as $e$ to obtain the prototype $\eta^e$. In summary, the objective function for calculating the prototype ${\eta}^e$ is designed as follows:
\begin{equation}
	{\eta}^{e}=\frac{\sum \big [F^{t-1}(X^t(i))*\mathbb{I}\big (e==\argmax \limits_{e' \in e_o\cup\mathcal{E}^{1:t-1}} \bm{\widehat{Y}}^{t-1}(i,e')\big ) \big ]}{\sum \mathbb{I}\big (e==\argmax \limits_{e' \in e_o\cup\mathcal{E}^{1:t-1}} \bm{\widehat{Y}}^{t-1}(i,e')\big )},
	\label{prototype}
\end{equation}
where $\mathbb{I}$ is the indicator function.

Finally, we briefly summarize \textbf{why the prototype is useful}. 
Firstly, the prototype approach treats all types equally, irrespective of their frequency of occurrence. This characteristic is particularly advantageous in tasks like NER, where the types may be unbalanced. Secondly, the prototype method is less sensitive to the outliers supposed to be the minority. Experimental results showcase that the proposed prototypical pseudo label strategy greatly enhances model performance by rectifying mistaken pseudo labels.

By implementing the prototypical pseudo label strategy, we obtain more high-quality pseudo labels for the old types, which offer robust semantic guidance in mitigating the background shift issue. The cross-entropy loss, computed using these constructed pseudo labels, is formulated as follows:
\begin{equation}
	\mathcal{L}_{\text{ce}}(\Theta^t)=-\frac{1}{|X^t|}\sum_{i=1}^{|X^t|} \bm{\widetilde{Y}}^t(i)\log\bm{\widehat{Y}}^t(i),
	\label{entropy_pseudo_label}
\end{equation}

\begin{table}[t]
  \centering
  \caption{The statistics for each NER dataset.}
  \resizebox{1.0\linewidth}{!}{
    \begin{tabular}{ccccc}
    \toprule
     \textbf{Dataset}     & \multicolumn{1}{l}{\# \textbf{Entity Type}} & \# \textbf{Sample}  & \multicolumn{2}{c}{\textbf{Entity Type Sequence (Alphabetical Order)}} \\
    \midrule
    CoNLL2003~\cite{sang1837introduction} & 4     & 21k    & \multicolumn{2}{l}{\begin{tabular}[1]{l}LOCATION, MISC, ORGANISATION, PERSON\end{tabular}} \\
    \midrule
    I2B2~\cite{murphy2010serving}  & 16    & 141k   & \multicolumn{2}{c}{\begin{tabular}[1]{l}AGE, CITY, COUNTRY, DATE, DOCTOR, HOSPITAL, \\  IDNUM, MEDICALRECORD, ORGANIZATION, \\PATIENT, PHONE, PROFESSION, STATE, STREET, \\USERNAME, ZIP\end{tabular}} \\
    \midrule
    OntoNotes5~\cite{hovy2006ontonotes} & 18    & 77k   & \multicolumn{2}{c}{\begin{tabular}[1]{l}CARDINAL, DATE, EVENT, FAC, GPE, LANGUAGE,\\ LAW, LOC, MONEY, NORP, ORDINAL, ORG,\\ PERCENT, PERSON, PRODUCT, QUANTITY, TIME,\\ WORK\_OF\_ART\end{tabular}} \\
    \bottomrule
    \end{tabular}%
    }
  \label{tab:dataset_statistics}%
\end{table}%

Overall, the objective function of our proposed RDP method can be formulated as follows:
\begin{equation}
	\mathcal{L}(\Theta^t)= \mathcal{L}_{\text{ce}}(\Theta^t)+\mathcal{L}_{\text{rd}}(\Theta^t).\\
	\label{gsc}
\end{equation}

\begin{table*}[t]
\centering
\caption{Comparisons with baselines under the FG-1-PG-1 setting of the CoNLL2003 dataset~\cite{sang1837introduction}. [LOC], [MISC], [ORG], and [PER] denote Location, All other types of entities, Organization, and Person. The \textcolor{beige}{\textbf{beige}} denotes the old entity types that have already been learned, and the \textcolor{orange}{\textbf{orange}} denotes the new entity types that are currently being learned. Mi. and Ma. represent Micro-F1 and Macro-F1 scores. The \textcolor{deepred}{\textbf{red}} denotes the highest result, and the \textcolor{blue}{\textbf{blue}} denotes the second highest result. The markers $\dagger$ refers to significant tests comparing with CFNER~\cite{zheng2022distilling} ($p$-$value$<$0.05$). $*$ represents results from our re-implementation. Other baseline results are directly cited from CFNER~\cite{zheng2022distilling}.}
\resizebox{1.0\linewidth}{!}{
\begin{tabular}{c|cc|cccc|cccc|cccc|>{\columncolor{lightgray}}c>{\columncolor{lightgray}}c}
		\toprule
		Task ID  & \multicolumn{2}{c|}{t=1 (base)} & \multicolumn{4}{c|}{t=2}&\multicolumn{4}{c|}{t=3} &\multicolumn{4}{c|}{t=4}& & \\
		Entity Type & \multicolumn{2}{c|}{All} &\multicolumn{1}{c|}{\textcolor{beige}{\textbf{[LOC]}}} & \multicolumn{1}{c|}{\textcolor{orange}{\textbf{[MISC]}}}&\multicolumn{2}{c|}{All} & \multicolumn{1}{c|}{\textcolor{beige}{\textbf{[LOC][MISC]}}} & \multicolumn{1}{c|}{\textcolor{orange}{\textbf{[ORG]}}}&\multicolumn{2}{c|}{All}& \multicolumn{1}{c|}{\textcolor{beige}{\textbf{[LOC][MISC][ORG]}}} & \multicolumn{1}{c|}{\textcolor{orange}{\textbf{[PER]}}}&\multicolumn{2}{c|}{All}& Avg.& Avg.  \\
        Evaluation Metric   & \multicolumn{1}{c}{Mi.}& \multicolumn{1}{c|}{Ma.}  & \multicolumn{1}{c|}{Ma.}  & \multicolumn{1}{c|}{Ma.} & \multicolumn{1}{c}{Mi.}& \multicolumn{1}{c|}{Ma.}  & \multicolumn{1}{c|}{Ma.} & \multicolumn{1}{c|}{Ma.}& \multicolumn{1}{c}{Mi.}& \multicolumn{1}{c|}{Ma.}  &
        \multicolumn{1}{c|}{Ma.} & \multicolumn{1}{c|}{Ma.}&\multicolumn{1}{c}{Mi.}& \multicolumn{1}{c|}{Ma.}  &

       Mi.& Ma. 
        \\
 \hline
         Only Finetuning  & 85.96±0.0 & 85.96±0.0 &\multicolumn{1}{c|}{--} &\multicolumn{1}{c|}{--} & 32.43±0.16&32.11±0.38  & \multicolumn{1}{c|}{--} & \multicolumn{1}{c|}{--} & 43.70±0.12 & 22.26±0.08 & \multicolumn{1}{c|}{--}  & \multicolumn{1}{c|}{--}& 41.27±0.12 & 22.22±0.16& 50.84±0.10 & 40.64±0.16  \\

         PODNet~\cite{DBLP:conf/eccv/DouillardCORV20}  & 85.96±0.0 & 85.96±0.0 &\multicolumn{1}{c|}{--} &\multicolumn{1}{c|}{--} & 11.13±0.67& 6.8±0.41 & \multicolumn{1}{c|}{--} & \multicolumn{1}{c|}{--} & 24.16±1.02 & 10.82±0.38 & \multicolumn{1}{c|}{--}  & \multicolumn{1}{c|}{--}& 25.49±0.37 & 13.95±0.31& 36.74±0.52 & 29.43±0.28  \\

     LUCIR~\cite{DBLP:conf/cvpr/HouPLWL19}  & 85.96±0.0 & 85.96±0.0 &\multicolumn{1}{c|}{--} &\multicolumn{1}{c|}{--} & 73.85±0.84& 70.47±0.69 &  \multicolumn{1}{c|}{--} & \multicolumn{1}{c|}{--} & 62.81±0.37 & 58.59±0.82 & \multicolumn{1}{c|}{--}  & \multicolumn{1}{c|}{--}& 73.78±0.51 & 66.68±1.12 & 74.15±0.43 & 70.48±0.66  \\

      Self-Training~\cite{DBLP:conf/wacv/RosenbergHS05}  & 85.96±0.0 & 85.96±0.0 &\multicolumn{1}{c|}{--} &\multicolumn{1}{c|}{--} & 74.96±2.11& 72.27±1.95  & \multicolumn{1}{c|}{--} & \multicolumn{1}{c|}{--} & 68.99±0.67 & 65.98±1.31 & \multicolumn{1}{c|}{--}  & \multicolumn{1}{c|}{--}& 74.79±0.84 & 67.32±1.22 & 76.17±0.91 & 72.88±1.12 \\

       $\text{ExtendNER}^{*}$~\cite{monaikul2021continual} & 85.96±0.0 & 85.96±0.0 &\multicolumn{1}{c|}{82.90±0.53} &\multicolumn{1}{c|}{58.89±2.45} & 73.76±1.27& 70.90±1.31  & \multicolumn{1}{c|}{66.36±1.34} & \multicolumn{1}{c|}{68.51±1.52} & 69.78±1.04 & 67.08±1.24 & \multicolumn{1}{c|}{62.00±1.53}  & \multicolumn{1}{c|}{88.11±1.59}& 75.00±0.62 & 68.53±1.13 & 76.07±0.35 & 73.06±0.29 \\

         ExtendNER~\cite{monaikul2021continual} & 85.96±0.0 & 85.96±0.0 &\multicolumn{1}{c|}{--} &\multicolumn{1}{c|}{--} & 74.42±0.99& 71.46±1.08  & \multicolumn{1}{c|}{--} & \multicolumn{1}{c|}{--} & 69.27±1.54 & 65.69±2.68 & \multicolumn{1}{c|}{--} & \multicolumn{1}{c|}{--}& 75.78±1.37 & 69.07±3.42 & 76.36±0.98 & 73.04±1.80 \\

 $\text{CFNER}^{*}$~\cite{zheng2022distilling} & 85.96±0.0 & 85.96±0.0 &\multicolumn{1}{c|}{85.62±0.79} &\multicolumn{1}{c|}{69.25±0.66} & 80.11±0.66& 77.44±0.60  & \multicolumn{1}{c|}{73.39±0.85} & \multicolumn{1}{c|}{74.48±0.45} & 75.11±0.25 & 73.71±0.49 & \multicolumn{1}{c|}{71.46±0.79}  & \multicolumn{1}{c|}{92.89±0.56}& 80.19±0.19 & 76.82±0.48 & 80.29±0.21 & 78.44±0.24\\

   CFNER~\cite{zheng2022distilling}  & 85.96±0.0 & \multicolumn{1}{c|}{85.96±0.0} &\multicolumn{1}{c|}{--} &\multicolumn{1}{c|}{--} & 80.63±0.54& \multicolumn{1}{c|}{78.19±0.56}  & \multicolumn{1}{c|}{--} & \multicolumn{1}{c|}{--} & 76.10±0.39 & \multicolumn{1}{c|}{74.95±0.47} & \multicolumn{1}{c|}{--}  & \multicolumn{1}{c|}{--}& 80.95±0.24 & \multicolumn{1}{c|}{77.34±0.96} & \textcolor{blue}{\textbf{80.91±0.29}}  & \textcolor{blue}{\textbf{79.11±0.50}} \\

 \hline

        \textbf{RDP (Ours)} & 85.96±0.0 & 85.96±0.0 &\multicolumn{1}{c|}{86.53±0.76} &\multicolumn{1}{c|}{73.01±1.45} & 82.49±0.85& 79.77±0.98  & \multicolumn{1}{c|}{76.79±0.39} & \multicolumn{1}{c|}{80.01±0.97} & 80.32±0.85 & 77.86±0.56 & \multicolumn{1}{c|}{76.29±1.00}  & \multicolumn{1}{c|}{94.53±0.34}& 83.34±1.00 & 80.85±0.73  &$\textcolor{deepred}{\textbf{82.55±0.26}}^{\dagger}$ &  $\textcolor{deepred}{\textbf{80.64±0.12}}^{\dagger}$ \\
     
  \textbf{Imp.}  & -- & -- &\multicolumn{1}{c|}{--} &\multicolumn{1}{c|}{--} &--& --  & \multicolumn{1}{c|}{--} & \multicolumn{1}{c|}{--} & -- & -- & \multicolumn{1}{c|}{--} & \multicolumn{1}{c|}{--}& -- & -- &  $\Uparrow$\textbf{1.64} &   $\Uparrow$\textbf{1.53}  \\

		\bottomrule
\end{tabular}
}
\label{tab:main_result_1_full}
\end{table*}

\begin{table*}[t]
\centering
\caption{Comparisons with baselines under the FG-2-PG-1 setting of the CoNLL2003 dataset~\cite{sang1837introduction}. [LOC], [MISC], [ORG], and [PER] denote Location, All other types of entities, Organization, and Person. The \textcolor{beige}{\textbf{beige}} denotes the old entity types that have already been learned, and the \textcolor{orange}{\textbf{orange}} denotes the new entity types that are currently being learned. Mi. and Ma. represent Micro-F1 and Macro-F1 scores. The \textcolor{deepred}{\textbf{red}} denotes the highest result, and the \textcolor{blue}{\textbf{blue}} denotes the second highest result. The markers $\dagger$ and $\ddagger$ refer to significant tests comparing with CFNER~\cite{zheng2022distilling} and LUCIR~\cite{DBLP:conf/cvpr/HouPLWL19} ($p$-$value$<$0.05$). $*$ represents results from our re-implementation. Other baseline results are directly cited from CFNER~\cite{zheng2022distilling}.}
\resizebox{1.0\linewidth}{!}{
\begin{tabular}{c|cc|cccc|cccc|>{\columncolor{lightgray}}c>{\columncolor{lightgray}}c}
		\toprule
		Task ID & \multicolumn{2}{c|}{t=1 (base)} & \multicolumn{4}{c|}{t=2}&\multicolumn{4}{c|}{t=3} & & \\
		Entity Type &\multicolumn{2}{c|}{All} &\multicolumn{1}{c|}{\textcolor{beige}{\textbf{[LOC][MISC]}}} & \multicolumn{1}{c|}{\textcolor{orange}{\textbf{[ORG]}}}&\multicolumn{2}{c|}{All} & \multicolumn{1}{c|}{\textcolor{beige}{\textbf{[LOC][MISC][ORG]}}} & \multicolumn{1}{c|}{\textcolor{orange}{\textbf{[PER]}}}&\multicolumn{2}{c|}{All}& Avg.& Avg.\\
  
        Evaluation Metric   & \multicolumn{1}{c}{Mi.}& \multicolumn{1}{c|}{Ma.}  & \multicolumn{1}{c|}{Ma.}  & \multicolumn{1}{c|}{Ma.}  & \multicolumn{1}{c}{Mi.}& \multicolumn{1}{c|}{Ma.}  &  \multicolumn{1}{c|}{Ma.}  & \multicolumn{1}{c|}{Ma.}  & \multicolumn{1}{c}{Mi.}& \multicolumn{1}{c|}{Ma.}  &
       Mi.& Ma. 
        \\
 \hline
        
	      Only Finetuning & 87.21±0.0 & \multicolumn{1}{c|}{84.33±0.0}&\multicolumn{1}{c|}{--}& \multicolumn{1}{c|}{--} & 43.73±0.07 & \multicolumn{1}{c|}{23.75±0.41} & \multicolumn{1}{c|}{--} &\multicolumn{1}{c|}{--}&41.41±0.09& \multicolumn{1}{c|}{22.66±0.14} & 57.45±0.05  & 43.58±0.18 \\

     PODNet~\cite{DBLP:conf/eccv/DouillardCORV20} &  87.21±0.0 & \multicolumn{1}{c|}{84.33±0.0}&\multicolumn{1}{c|}{--}& \multicolumn{1}{c|}{--} & 46.14±0.39 & \multicolumn{1}{c|}{48.13±0.74} & \multicolumn{1}{c|}{--} &\multicolumn{1}{c|}{--}&44.24±1.24& \multicolumn{1}{c|}{42.89±2.23} & 59.12±0.54 & 58.39±0.99\\

     LUCIR~\cite{DBLP:conf/cvpr/HouPLWL19} &  87.21±0.0 & \multicolumn{1}{c|}{84.33±0.0}&\multicolumn{1}{c|}{--}& \multicolumn{1}{c|}{--} & 74.59±0.71 & \multicolumn{1}{c|}{73.46±0.69} & \multicolumn{1}{c|}{--} &\multicolumn{1}{c|}{--}&80.02±0.23& \multicolumn{1}{c|}{74.36±0.24} & 80.53±0.31  & \textcolor{blue}{\textbf{77.33±0.31}}\\

      Self-Training~\cite{DBLP:conf/wacv/RosenbergHS05} & 87.21±0.0 & \multicolumn{1}{c|}{84.33±0.0}&\multicolumn{1}{c|}{--}& \multicolumn{1}{c|}{--} & 67.34±0.34 & \multicolumn{1}{c|}{51.12±0.07} & \multicolumn{1}{c|}{--} &\multicolumn{1}{c|}{--}&75.41±0.37& \multicolumn{1}{c|}{64.71±0.26} & 76.65±0.24 & 66.72±0.11 \\
      
       $\text{ExtendNER}^{*}$~\cite{monaikul2021continual}&  87.21±0.0 & \multicolumn{1}{c|}{84.33±0.0}&\multicolumn{1}{c|}{52.59±2.34}& \multicolumn{1}{c|}{66.20±1.03} & 69.63±0.88 & \multicolumn{1}{c|}{57.12±1.84} & \multicolumn{1}{c|}{60.08±2.56} &\multicolumn{1}{c|}{92.48±0.62}&76.69±0.75& \multicolumn{1}{c|}{68.18±1.91} & 77.89±0.42 & 69.92±1.02 \\

         ExtendNER~\cite{monaikul2021continual}& 87.21±0.0 & \multicolumn{1}{c|}{84.33±0.0}&\multicolumn{1}{c|}{--}& \multicolumn{1}{c|}{--} & 67.93±0.88 & \multicolumn{1}{c|}{51.92±0.78} & \multicolumn{1}{c|}{--} &\multicolumn{1}{c|}{--}&74.84±1.11& \multicolumn{1}{c|}{62.84±1.15} & 76.66±0.66 & 66.36±0.64 \\

 $\text{CFNER}^{*}$~\cite{zheng2022distilling} & 87.21±0.0 & \multicolumn{1}{c|}{84.33±0.0}&\multicolumn{1}{c|}{71.68±1.38}& \multicolumn{1}{c|}{74.25±0.59} & 77.09±0.69 & \multicolumn{1}{c|}{72.54±1.10} & \multicolumn{1}{c|}{68.20±1.94} &\multicolumn{1}{c|}{93.90±0.33}&80.11±0.63& \multicolumn{1}{c|}{74.62±1.39} & \textcolor{blue}{\textbf{81.52±0.43}} & 77.20±0.82 \\

   CFNER~\cite{zheng2022distilling} & 87.21±0.0 & \multicolumn{1}{c|}{84.33±0.0}&\multicolumn{1}{c|}{--}& \multicolumn{1}{c|}{--} & 76.23±0.36 & \multicolumn{1}{c|}{69.19±0.28} & \multicolumn{1}{c|}{--} &\multicolumn{1}{c|}{--}&79.05±0.73& \multicolumn{1}{c|}{72.09±0.67} & 80.83±0.36 & 75.20±0.32 \\

 \hline

        \textbf{RDP (Ours)} &  87.21±0.0 & \multicolumn{1}{c|}{84.33±0.0}&\multicolumn{1}{c|}{83.23±0.68}& \multicolumn{1}{c|}{82.39±0.54} & 84.70±0.51 & \multicolumn{1}{c|}{82.95±0.56} & \multicolumn{1}{c|}{81.65±0.90} &\multicolumn{1}{c|}{94.28±0.48}&86.86±0.70& \multicolumn{1}{c|}{84.81±0.66} & $\textcolor{deepred}{\textbf{85.82±0.36}}^{\dagger}$ &  $\textcolor{deepred}{\textbf{83.59±0.37}}^{\ddagger}$ \\

  \textbf{Imp.} & -- & -- &\multicolumn{1}{c|}{--}&\multicolumn{1}{c|}{--}&--&--&\multicolumn{1}{c|}{--}&\multicolumn{1}{c|}{--}&--&--& $\Uparrow$\textbf{4.30} &  $\Uparrow$\textbf{6.26}\\

		\bottomrule
\end{tabular}
}
\label{tab:main_result_2_full}
\end{table*}

\section{Experimental Setup}

To ensure a fair comparison with SOTA methods, we adhere to the experimental setup of CFNER~\cite{zheng2022distilling}. This includes using the same benchmark datasets, INER settings, competing baselines, evaluation metrics, and basic implementation details.

\subsection{Benckmark Datasets}

We evaluate our RDP on three widely used NER datasets: CoNLL2003 \cite{sang1837introduction}, I2B2~\cite{murphy2010serving}, and OntoNotes5~\cite{hovy2006ontonotes}. The dataset statistics are summarized in Table~\ref{tab:dataset_statistics}.

We employ the greedy sampling algorithm proposed in CFNER \cite{zheng2022distilling} to split the training set into disjoint slices (corresponding to different incremental learning tasks). 
This algorithm ensures that the samples of each entity type are predominantly distributed in the corresponding slice, simulating real-world scenarios more effectively. 
In each slice, we only retain labels belonging to the entity types to be learned while masking other labels as the non-entity type. For further details on the greedy sampling algorithm, please refer to Appendix B of CFNER~\cite{zheng2022distilling}.

\begin{table*}[t]
\centering
\caption{Comparisons with baselines on the I2B2~\cite{murphy2010serving} and OntoNotes5~\cite{hovy2006ontonotes} datasets. The \textcolor{deepred}{\textbf{red}} denotes the highest result, and the \textcolor{blue}{\textbf{blue}} denotes the second highest result. The marker $\dagger$ refers to significant test $p$-$value$<$0.05$ comparing with CFNER~\cite{zheng2022distilling}. $*$ represents results from our re-implementation. Other baseline results are directly cited from CFNER~\cite{zheng2022distilling}.}
\resizebox{1.0\linewidth}{!}{%
\begin{tabular}{@{}cccccccccc@{}}
\toprule
 &  & \multicolumn{2}{c}{FG-1-PG-1} & \multicolumn{2}{c}{FG-2-PG-2} & \multicolumn{2}{c}{FG-8-PG-1} & \multicolumn{2}{c}{FG-8-PG-2} \\ \cmidrule(l){3-10} 
\multirow{-2}{*}{Dataset} & \multirow{-2}{*}{Baseline} & Micro-F1  & Macro-F1  & Micro-F1  & Macro-F1  & Micro-F1  & Macro-F1  & Micro-F1  & Macro-F1  \\ \midrule
 & Only Finetuning & 17.43±0.54 & 13.81±1.14 &  28.57±0.26 & 21.43±0.41 &  20.83±1.78 & 18.11±1.66 &  23.60±0.15 & 23.54±0.38\\
 
 & PODNet~\cite{DBLP:conf/eccv/DouillardCORV20} & 12.31±0.35 & 17.14±1.03 & 34.67±2.65 & 24.62±1.76& 39.26±1.38 & 27.23±0.93 & 36.22±12.9 & 26.08±7.42\\

 & LUCIR~\cite{DBLP:conf/cvpr/HouPLWL19} & 43.86±2.43 & 31.31±1.62 &  64.32±0.76 & 43.53±0.59 &  57.86±0.87 & 33.04±0.39 &  68.54±0.27 & 46.94±0.63\\

 & Self-Training~\cite{DBLP:conf/wacv/RosenbergHS05} & 31.98±2.12 & 14.76±1.31 & 55.44±4.78 & 33.38±3.13 & 49.51±1.35 & 23.77±1.01 & 48.94±6.78 & 29.00±3.04\\

  & $\text{ExtendNER}^{*}$~\cite{monaikul2021continual} & 41.65±10.11 & 23.11±2.70 &  67.60±1.15 &  42.58±1.59 &  45.14±2.91 &  27.41±0.88 &  56.48±2.41 & 38.88±1.38 \\

 & ExtendNER~\cite{monaikul2021continual} & 42.85±2.86 & 24.05±1.35 & 57.01±4.14 & 35.29±3.38 & 43.95±2.01 & 23.12±1.79 & 52.25±5.36  & 30.93±2.77 \\

 & $\text{CFNER}^{*}$~\cite{zheng2022distilling} & \textcolor{blue}{\textbf{64.79±0.26}} & \textcolor{blue}{\textbf{37.79±0.65}} & \textcolor{blue}{\textbf{72.58±0.59}} & \textcolor{blue}{\textbf{51.71±0.84}} & 56.66±3.22 & 36.84±1.35 & \textcolor{blue}{\textbf{69.12±0.94}} & \textcolor{blue}{\textbf{51.61±0.87}} \\

 \multirow{-6}{*}{I2B2~\cite{murphy2010serving}} & CFNER~\cite{zheng2022distilling} & 62.73±3.62 & 36.26±2.24 & 71.98±0.50 & 49.09±1.38 & \textcolor{blue}{\textbf{59.79±1.70}} & \textcolor{blue}{\textbf{37.30±1.15}} & 69.07±0.89 & 51.09±1.05 \\

\cmidrule(l){2-10}
 & \textbf{RDP (Ours)} & $\textcolor{deepred}{\textbf{71.39±1.01}}^{\dagger}$ & $\textcolor{deepred}{\textbf{44.00±2.31}}^{\dagger}$ & $\textcolor{deepred}{\textbf{77.45±0.55}}^{\dagger}$ & $\textcolor{deepred}{\textbf{53.48±0.66}}^{\dagger}$ & $\textcolor{deepred}{\textbf{77.50±1.26}}^{\dagger}$& $\textcolor{deepred}{\textbf{62.99±0.36}}^{\dagger}$& $\textcolor{deepred}{\textbf{80.08±0.40}}^{\dagger}$ & $\textcolor{deepred}{\textbf{63.72±0.71}}^{\dagger}$\\

  & \textbf{Imp.} &  $\Uparrow$\textbf{6.60} &   $\Uparrow$\textbf{6.21} &   $\Uparrow$\textbf{4.87} &  $\Uparrow$\textbf{1.77} & $\Uparrow$\textbf{17.71} &   $\Uparrow$\textbf{25.69} &   $\Uparrow$\textbf{10.96} &  $\Uparrow$\textbf{12.11}\\

\midrule
\midrule

 & Only Finetuning & 15.27±0.26 & 10.85±1.11 &  25.85±0.11 & 20.55±0.24 &  17.63±0.57 & 12.23±1.08 &  29.81±0.12 & 20.05±0.16\\
 
 & PODNet~\cite{DBLP:conf/eccv/DouillardCORV20} & 9.06±0.56 & 8.36±0.57 & 34.67±1.08 & 24.62±0.85 & 29.00±0.86 & 20.54±0.91 & 37.38±0.26 & 25.85±0.29\\

 & LUCIR~\cite{DBLP:conf/cvpr/HouPLWL19} & 28.18±1.15 & 21.11±0.84 &  64.32±1.79 & 43.53±1.11 &  66.46±0.46 & 46.29±0.38 &  76.17±0.09 & 55.58±0.55\\

 & Self-Training~\cite{DBLP:conf/wacv/RosenbergHS05} & 50.71±0.79 & 33.24±1.06 & 68.93±1.67 & 50.63±1.66 & 73.59±0.66 & 49.41±0.77 & 77.07±0.62 & 53.32±0.63\\

 & $\text{ExtendNER}^*$~\cite{monaikul2021continual} & 51.36±0.77 & 33.38±0.98 & 63.03±9.39 & 47.64±5.15 & 73.65±0.19 & 50.55±0.56 & 77.86±0.10  & 55.21±0.51 \\

  & ExtendNER~\cite{monaikul2021continual} & 50.53±0.86 & 32.84±0.84 & 67.61±1.53 & 49.26±1.49 & 73.12±0.93 & 49.55±0.90 & 76.85±0.77  & 54.37±0.57 \\

 & $\text{CFNER}^*$~\cite{zheng2022distilling} & 58.44±0.71 & 41.75±1.51 & 
 72.10±0.31 & 
 55.02±0.35 & 
 78.25±0.33 &
 \textcolor{blue}{\textbf{58.64±0.42}} & 
80.09±0.37 & 
 \textcolor{blue}{\textbf{61.06±0.37}} \\

 \multirow{-6}{*}{OntoNotes5~\cite{hovy2006ontonotes}} & CFNER~\cite{zheng2022distilling} & \textcolor{blue}{\textbf{58.94±0.57}} & \textcolor{blue}{\textbf{42.22±1.10}} & \textcolor{blue}{\textbf{72.59±0.48}} & \textcolor{blue}{\textbf{55.96±0.69}} & \textcolor{blue}{\textbf{78.92±0.58}} & 57.51±1.32 &  \textcolor{blue}{\textbf{80.68±0.25}} & 60.52±0.84 \\

\cmidrule(l){2-10}
 & \textbf{RDP (Ours)} & $\textcolor{deepred}{\textbf{68.28±1.09}}^{\dagger}$ & $\textcolor{deepred}{\textbf{53.56±0.39}}^{\dagger}$ & $\textcolor{deepred}{\textbf{74.38±0.26}}^{\dagger}$ & $\textcolor{deepred}{\textbf{57.73±0.54}}^{\dagger}$ & $\textcolor{deepred}{\textbf{79.89±0.20}}^{\dagger}$& $\textcolor{deepred}{\textbf{63.20±0.58}}^{\dagger}$& $\textcolor{deepred}{\textbf{83.30±0.30}}^{\dagger}$ & $\textcolor{deepred}{\textbf{66.92±1.26}}^{\dagger}$\\

   & \textbf{Imp.} &  $\Uparrow$\textbf{9.34} &   $\Uparrow$\textbf{11.34} &   $\Uparrow$\textbf{1.79} &  $\Uparrow$\textbf{1.77} & $\Uparrow$\textbf{0.97} &   $\Uparrow$\textbf{4.56} &   $\Uparrow$\textbf{2.62} &  $\Uparrow$\textbf{5.86}\\

\bottomrule
\end{tabular}
}
\label{tab:main_result_3_full}
\end{table*}

\begin{figure*}[tbp]
\centering
  \includegraphics[width=1.0\linewidth]{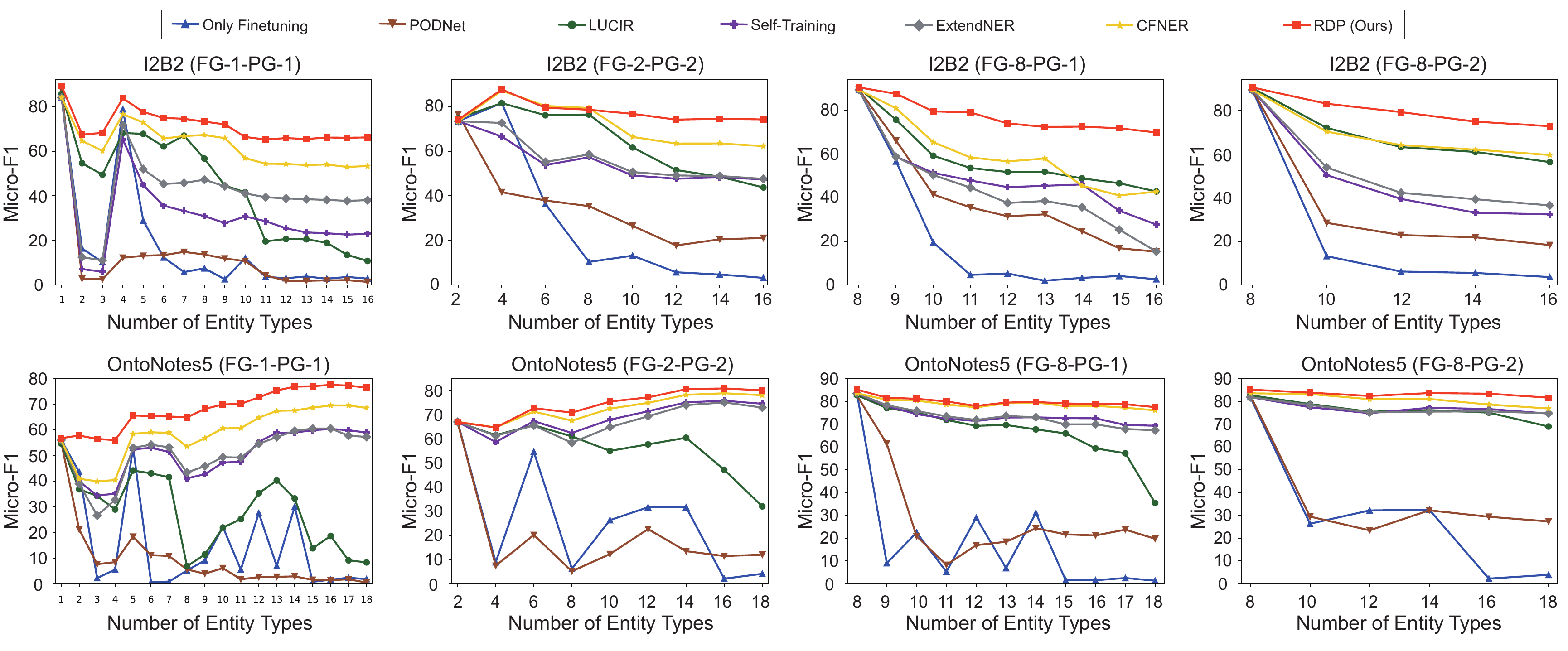}
    \caption{Comparison of the task-wise Micro-F1 on I2B2~\cite{murphy2010serving} and OntoNotes5~\cite{hovy2006ontonotes}. Results of baselines are from CFNER~\cite{zheng2022distilling}.}
    \label{fig:main_result_step_3_mi}
\end{figure*}

\subsection{INER Settings}

Following CFNER~\cite{zheng2022distilling}, we consider two different INER scenarios for each dataset. The first scenario involves training the base model (the first task) using the same number of entity types as the subsequent incremental learning tasks. The second scenario trains the base model with half of all entity types. 
The former is more challenging, while the latter is closer to real-world situations, allowing models to acquire sufficient knowledge before incremental learning.
During training, we learn entity types in a fixed order, specifically in alphabetical order, and use the corresponding data slice to train models sequentially. 
The FG entity types are used to train the base model, and in each incremental learning task, the PG entity types are introduced, denoted as FG-a-PG-b. 
For the CoNLL2003 dataset~\cite{sang1837introduction}, we set two settings: FG-1-PG-1 and FG-2-PG-1. 
For the I2B2~\cite{murphy2010serving} and OntoNotes5~\cite{hovy2006ontonotes} datasets, we set four settings: FG-1-PG-1, FG-2-PG-2, FG-8-PG-1, and FG-8-PG-2.
During the evaluation, we retain only the labels of the current entity types in the validation set, setting other labels as the non-entity type. In each incremental learning task, we select the model with the best validation performance for testing and the subsequent learning task. 
For testing, we retain labels for all learned entity types while setting others as the non-entity type in the test set.

\subsection{Competing Baselines}

We evaluate the performance of our RDP by comparing it with recent INER methods, namely ExtendNER~\cite{monaikul2021continual} and CFNER~\cite{zheng2022distilling}, where CFNER represents the previous SOTA method. Additionally, we include the lower bound comparison of Only Finetuning, which does not employ any anti-forgetting technique.
Furthermore, we compare our method with incremental learning methods used in computer vision, such as Self-Training~\cite{DBLP:conf/wacv/RosenbergHS05,DBLP:journals/corr/abs-1909-08383}, LUCIR~\cite{DBLP:conf/cvpr/HouPLWL19}, and PODNet~\cite{DBLP:conf/eccv/DouillardCORV20}. It's worth mentioning that Zheng et al.~\cite{zheng2022distilling} adapted these methods to the INER scenario. For a more detailed introduction of the baselines, please refer to Appendix C of CFNER~\cite{zheng2022distilling}.

\subsection{Evaluation Metrics and Significance Test}
In line with CFNER~\cite{zheng2022distilling}, we employ the Micro and Macro F1 scores as evaluation metrics to account for the entity type imbalance issue in NER and assess the performance of our method. The final performance is reported as the average across all incremental learning tasks, including the first task.
Moreover, we present task-wise performance comparison line plots to provide a finer-grained analysis. To evaluate the statistical significance of the improvements, we conduct a paired t-test with a significance level of $0.05$~\cite{koehn2004statistical}.

\subsection{Implementation Details}

Following the previous INER methods~\cite{monaikul2021continual,xia2022learn,zheng2022distilling}, we utilize the "BIO" tagging schema for all three datasets. This schema assigns two labels to each entity type: B-type (for the beginning of an entity) and I-type (for the inside of an entity).
Our NER model employs the bert-base-cased~\cite{kenton2019bert} model as the encoder and utilizes a fully-connected layer as the classifier. The model is implemented in the PyTorch framework~\cite{paszke2019pytorch} on top of the BERT Huggingface implementation~\cite{wolf2019huggingface}.
Consistent with CFNER~\cite{zheng2022distilling}, we train the model for $20$ epochs if PG=$2$, and $10$ epochs otherwise. The learning rate, batch size, and trade-off hyperparameters $\lambda_1$ and $\lambda_2$ are set to $4e$-$4$, $8$, $0.3$, and $0.1$, respectively.
All experiments are conducted on a single NVIDIA A100 GPU with $40$GB of memory, and each experiment is run $5$ times to ensure statistical robustness.

\section{Results and Discussions}

To validate the superiority and effectiveness of our RDP method, we conducted extensive experiments to answer the following research questions (RQ):

\begin{itemize}
\item \textbf{RQ1}: What is the quantitative performance of RDP compared to competitive baselines?
\item \textbf{RQ2}: How about the contribution of the critical components in RDP? 
\item \textbf{RQ3}: What is the qualitative performance of RDP compared to competitive baselines?
\end{itemize}

\subsection{Main Results (RQ1)}

In this subsection, we conducted extensive experiments on the CoNLL2003~\cite{sang1837introduction}, I2B2~\cite{murphy2010serving}, and OntoNotes5~\cite{hovy2006ontonotes} datasets under ten INER settings to demonstrate the quantitative performance of our RDP compared to competitive baselines. The results are presented in Tables~\ref{tab:main_result_1_full}, \ref{tab:main_result_2_full}, \ref{tab:main_result_3_full}, and Figure~\ref{fig:main_result_step_3_mi}.
For the two INER settings of the CoNLL2003 dataset~\cite{sang1837introduction}, FG-1-PG-1 and FG-2-PG-1, detailed metrics are provided in Tables~\ref{tab:main_result_1_full} and \ref{tab:main_result_2_full}, respectively. These metrics include separate Macro F1 scores for old and new types in each incremental task, as well as Micro and Macro F1 scores for all types within each task, and the final average Micro and Macro F1 scores across all tasks.
Due to space limitations, only the final average Micro and Macro F1 scores across all tasks are provided in Table~\ref{tab:main_result_3_full} for the eight INER settings in the I2B2~\cite{murphy2010serving} and OntoNotes5~\cite{hovy2006ontonotes} datasets.
As a supplement, line plots depicting task-wise Micro F1 score comparisons (Micro F1 scores for all types in each task) for the eight INER settings in the I2B2~\cite{murphy2010serving} and OntoNotes5~\cite{hovy2006ontonotes} datasets are shown in Figure~\ref{fig:main_result_step_3_mi}. However, line plots for task-wise Macro F1 score comparisons are not provided due to space limitations. Nevertheless, the observations for task-wise Macro F1 score comparisons are similar to those observed for task-wise Micro F1 score comparisons.

As shown in Table~\ref{tab:main_result_1_full} and Table~\ref{tab:main_result_2_full}, our RDP demonstrates improvements over the previous SOTA baselines LUCIR~\cite{DBLP:conf/cvpr/HouPLWL19} and CFNER~\cite{zheng2022distilling} by approximately 1.64\% and 4.30\% in Micro-F1, and 1.53\% and 6.26\% in Macro-F1, under two INER settings (FG-1-PG-1 and FG-2-PG-1) of the CoNLL2003 dataset~\cite{sang1837introduction}.
As depicted in the upper part of Table~\ref{tab:main_result_3_full}, our RDP achieves improvements over the previous SOTA baseline CFNER~\cite{zheng2022distilling} ranging from 4.87\% to 17.71\% in Micro-F1, and 1.77\% to 25.69\% in Macro-F1, under four INER settings (FG-1-PG-1, FG-2-PG-2, FG-8-PG-1, and FG-8-PG-2) of the I2B2 dataset~\cite{murphy2010serving}. Similarly, in the lower part of Table~\ref{tab:main_result_3_full}, our RDP achieves improvements over the previous SOTA baseline CFNER~\cite{zheng2022distilling} ranging from 0.97\% to 9.34\% in Micro-F1, and 1.77\% to 11.34\% in Macro-F1, under four INER settings (FG-1-PG-1, FG-2-PG-2, FG-8-PG-1, and FG-8-PG-2) of the OntoNotes5 dataset~\cite{hovy2006ontonotes}.
Furthermore, as illustrated in Figure~\ref{fig:main_result_step_3_mi}, our RDP outperforms the INER baselines in task-wise Micro-F1 comparisons across the eight settings of the I2B2~\cite{murphy2010serving} and OntoNotes5~\cite{hovy2006ontonotes} datasets. These results quantitatively confirm the superiority and effectiveness of our RDP compared to competitive baselines, showcasing its ability to learn a robust INER model and indicating improved resilience to catastrophic forgetting and background shift problems.

\begin{table}[tbp]
  \centering
  \caption{The ablation study of our RDP under the FG-1-PG-1 setting of the I2B2~\cite{murphy2010serving} and OntoNotes5~\cite{hovy2006ontonotes} datasets. Compared with our RDP, all ablation variants significantly degrade INER performance, verifying the importance of all components to address INER collaboratively.}
  \resizebox{1.0\linewidth}{!}{
    \begin{tabular}{lcccc}
    \toprule
    \multirow{2}[4]{*}{Method} & \multicolumn{2}{c}{I2B2} & \multicolumn{2}{c}{OntoNotes5}  \\
\cmidrule{2-5}          & Micro-F1 & Macro-F1 & Micro-F1 & Macro-F1  \\
    \midrule
    
    \textbf{RDP (Ours)}  & \textbf{71.39±1.01}  & \textbf{44.00±2.31}  & \textbf{68.28±1.09}  & \textbf{53.56±0.39}  \\

    \ \ \ \ \ \ w/o $\mathcal{L}_\text{cd}$ &64.97±0.55& 38.76±1.01& 63.56±0.37& 47.49±1.36\\
     \ \ \ \ \ \ w/o $\mathcal{L}_\text{se}$ &  67.59±1.42 &41.32±2.66 & 65.47±0.43 & 50.27±0.59\\
      \ \ \ \ \ \ w/o PPL &  64.17±1.19 &39.86±2.03 & 64.09±0.57 & 46.09±0.80\\
     \ \ \ \ \ \ w/o PL &  48.93±0.69 &31.66±0.64 & 56.64±0.45 & 39.54±1.05\\
    \bottomrule
    \end{tabular}
    }
  \label{tab:ablation_study}
\end{table}

\begin{figure*}[t]
\centering
  \includegraphics[width=1.0\linewidth]{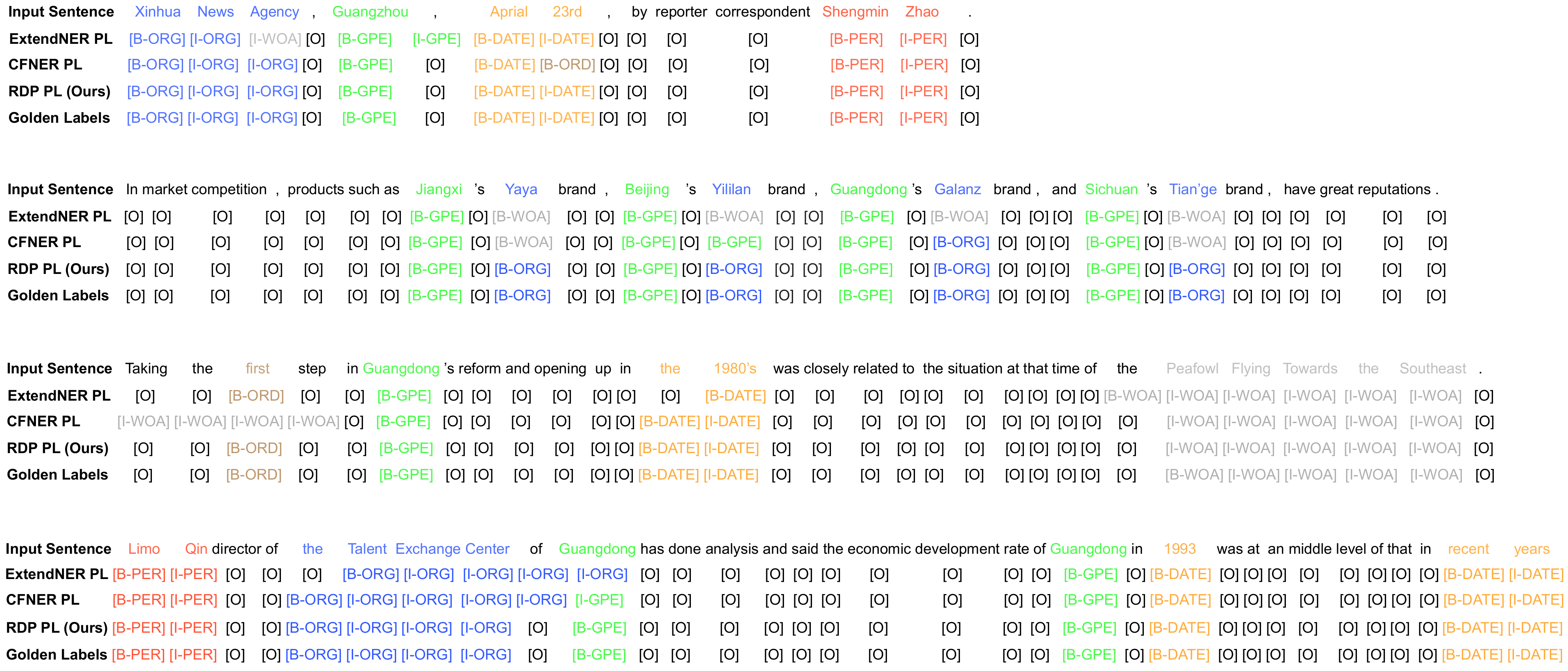}
  \caption{Four real NER cases sampled from the OntoNotes5~\cite{hovy2006ontonotes} test set. \textbf{PL} denotes the predicted labels. \textbf{B-} and \textbf{I-} distinguish begin/inside of named entities. [O], \textcolor{orange}{[DATE]}, \textcolor{green}{[GPE]}, \textcolor{brown}{[ORD]}, \textcolor{blue}{[ORG]}, \textcolor{red}{[PER]}, and \textcolor{gray}{[WOA]} denote non-entity type, \textcolor{orange}{\texttt{Date}}, \textcolor{green}{\texttt{Countries, Cities, or States}}, \textcolor{brown}{\texttt{Ordinals}}, \textcolor{blue}{\texttt{Organization}}, \textcolor{red}{\texttt{Person}}, and \textcolor{gray}{\texttt{Work of art}}, respectively. All the prediction results are from the last task of the FG-8-PG-2 setting. These visualization NER cases qualitatively demonstrate the superiority and effectiveness of our proposed RDP method.}
\label{fig:case}
\end{figure*}

\subsection{Ablation Study (RQ2)}

In this subsection, we conducted ablation studies to analyze the effects of critical components in our RDP, as presented in Table~\ref{tab:ablation_study}.
The results from \textbf{RDP} w/o $\mathcal{L}{\text{cd}}$ and w/o $\mathcal{L}{\text{se}}$ indicate the performance of our method without inter-task relation distillation loss and intra-task self-entropy loss. These results demonstrate the essential roles played by both $\mathcal{L}{\text{cd}}$ and $\mathcal{L}{\text{se}}$ in our method. The inter-task relation distillation loss ensures semantic consistency across different incremental learning tasks, while the intra-task self-entropy loss increases prediction confidence, striking a suitable stability-plasticity trade-off and effectively mitigating catastrophic forgetting.
\textbf{RDP} w/o PPL refers to using the max output probability of the old model directly (naive pseudo label strategy) without utilizing our proposed prototypical pseudo label strategy. On the other hand, \textbf{RDP} w/o PL indicates that no pseudo label strategy is employed. The results highlight that incorporating pseudo labels allows the NER model to leverage the old entity type information contained in non-entity type tokens and recall what it has learned from them. Our prototypical pseudo label strategy effectively rectifies prediction errors from the old model by reweighting the old output probability distribution based on the distances between token embeddings and type-wise prototypes. This approach better addresses background shift by using high-quality pseudo labels.
Overall, the ablation studies affirm the significance of the critical components in our RDP, showcasing their contributions to improving the model's performance and addressing challenges such as catastrophic forgetting and background shift.

\subsection{Case Study (RQ3)}

In this subsection, we showcase the qualitative performance of our RDP in comparison to competitive baselines, including ExtendNER~\cite{monaikul2021continual} and CFNER~\cite{zheng2022distilling}. We present visualized qualitative comparison results from the FG-8-PG-2 setting on the OntoNotes5 dataset~\cite{hovy2006ontonotes} in Figure~\ref{fig:case}.
These qualitative comparisons emphasize the superiority of our RDP over the examined baselines in terms of learning new entity types consecutively, reinforcing its effectiveness in the incremental learning scenario.

\section{Conclusion}

In this paper, we present the RDP method as a solution to address the challenges of catastrophic forgetting and background shift in INER. 
We begin by introducing a task relation distillation scheme to explore the semantic relations between old and new tasks, leading to a suitable trade-off between stability and plasticity for INER and, ultimately, mitigating catastrophic forgetting. 
Additionally, we propose a prototypical pseudo label strategy to label old entity types contained in the non-entity type, effectively tackling the background shift problem by correcting the prediction error from the old model. 
We conduct extensive experiments on ten INER settings of three datasets: CoNLL2003, I2B2, and OntoNotes5. 
The results clearly show the superiority of our RDP method, outperforming previous SOTA methods by a significant margin. Our method offers a promising direction for advancing INER techniques and overcoming the challenges posed by incremental learning scenarios.

\bibliographystyle{ACM-Reference-Format}
\balance
\bibliography{sample-base}

\appendix

\end{document}